\pdfoutput=1

\documentclass[11pt]{article}

\usepackage[final]{acl}

\usepackage{times}
\usepackage{latexsym}
\usepackage{graphicx}
\usepackage{url}
\usepackage{algorithm}
\usepackage{algpseudocode}
\usepackage{amsmath} 
\usepackage{booktabs}

\usepackage[T1]{fontenc}

\usepackage[utf8]{inputenc}

\usepackage{microtype}

\usepackage{inconsolata}

\title{Recursive Chain-of-Feedback Prevents Performance Degradation from Redundant Prompting}

\author{Jinwoo Ahn\thanks{Corresponding Author} \hspace{0.5cm} Kyuseung Shin \\
  University of California, Berkeley \\
\texttt{\{andrewahn, kyuds1016\}@berkeley.edu}}

\begin{document}
\maketitle
\begin{abstract}
Large Language Models (LLMs) frequently struggle with complex reasoning tasks, failing to construct logically sound steps towards the solution. In response to this behavior, users often try prompting the LLMs repeatedly in hopes of reaching a better response. This paper studies such repetitive behavior and its effect by defining a novel setting, \textbf{C}hain-\textbf{o}f-\textbf{F}eedback (CoF). The setting takes questions that require multi-step reasoning as an input. Upon response, we repetitively prompt meaningless feedback (e.g. '\textit{make another attempt}') requesting additional trials. Surprisingly, our preliminary results show that repeated meaningless feedback gradually decreases the quality of the responses, eventually leading to a larger deviation from the intended outcome. To alleviate these troubles, we propose a novel method, \textbf{R}ecursive \textbf{C}hain-\textbf{o}f-\textbf{F}eedback (R-CoF). Following the logic of recursion in computer science, R-CoF recursively revises the initially incorrect response by breaking down each incorrect reasoning step into smaller individual problems. Our preliminary results show that majority of questions that LLMs fail to respond correctly can be answered using R-CoF without any sample data outlining the logical process.
\end{abstract}
\section{Introduction}
Relying heavily on AI agents can have profound risks in many different contexts. Nevertheless, growing literature has shown that the accessibility and capability of public conversational language models (e.g. ChatGPT) have caused the general public to take these models as a reliable source of information \citep{peter2023advice}, often as the ground truth despite the lack of a concrete verification method \citep{doyou2023trust}. For instance, suppose a group of students attempts to solve a series of strenuous math questions but without access to the solutions. To verify the answers to the questions, students might input the problem statements into ChatGPT. If the answer seems reasonable, they make a naive assumption that the provided response could serve as the solution. Otherwise, they will repetitively prompt the LLMs for another attempt, often with increasing negative sentiment.
\begin{figure}[t!]
    \centering
    \includegraphics[width=1\linewidth]{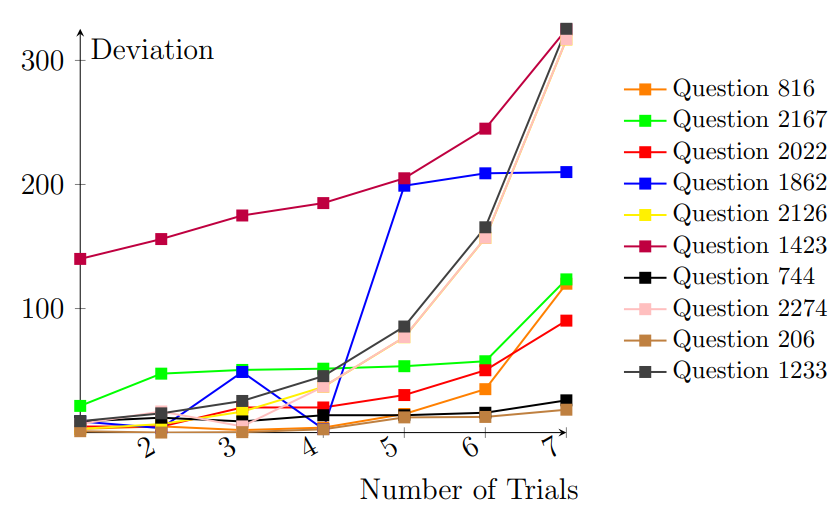}
    \caption{A sample of experimental results are shown here, where the deviation continues to increase as the number of trials the LLM takes to revise its answer from meaningless feedback increases.
    }
    \label{fig:graph}
\end{figure}

Recent studies, however, have shown that LLMs are prone to generate contradicting sentences \citep{niels2023contradict} or be distracted with irrelevant context \citep{freda2023distract}, ultimately leading to hallucination. To prevent these problems, many new prompt engineering methods, often variations of Chain-of-Thought prompting \citep{jason2022cot}, have been introduced \citep{openai2022, verify2023edit, code2023chain}. Despite the promising results presented in these studies, these prompting methods still suffer from requiring a sample question, reasoning steps, and solution that closely resemble the target question. When faced with complicated reasoning problems that individuals suffer to even begin with, these prompting methods are often not the most viable approach. As such, the discovery of a prompt engineering method that increases the reliability of LLM responses without the need for a similar sample point (i.e. zero-shot) still remains an unresolved task.

In order to raise awareness of the unreliable behavior of LLMs, we first formulate a new setting: \textbf{C}hain-\textbf{o}f-\textbf{F}eedback (CoF). The setting takes two stages of input-output sequence. First, it receives a multi-step reasoning question and returns its initial response. For this version of the paper, we only perform preliminary experiments with the input to the CoF setting restricted to multi-step math questions with definite answers. Then, it takes a meaningless feedback as an input (e.g. '\textit{make another attempt}') requesting additional trials. Here, we define 'meaningless' to meet the following criteria: not providing 1) additional clarifications about the question or 2) guidance towards the correct direction. Then, we randomly sample a question that the LLM failed to provide the correct output initially, request the LLM to correct its response through meaningless feedback, and calculate the absolute deviation from the correct answer. Our preliminary results show that the responses diverge more with increasing numbers of iteration. This behavior provides insight towards the idea that meaningless feedback makes the response worse.

Then, to tackle the problem of verifying the model responses, we present the following novel method: \textbf{R}ecursive \textbf{C}hain \textbf{o}f \textbf{F}eedback (R-CoF). Similar to the CoF setting, R-CoF takes in a multi-step reasoning question. Then, it outputs a response and the reasoning steps it took to reach the final answer. Given that the initial response is incorrect, the LLM freezes all the correct steps, makes a recursive call to a separate LLM to adjust the incorrect step, and finally incorporates the correct reasoning in the process of refining its final answer. By requesting a separate LLM without the knowledge of the original question to attempt a smaller portion of it, R-CoF essentially follows the process of recursion in computer science and increases the chances of adjusting the incorrect reasoning step.

In summary, aside from the quantitative benefits, we make the following qualitative contributions by introducing the CoF setting and R-CoF method:
\begin{itemize}
    \item \textbf{Raise awareness of the practices that trigger more divergence.} Past works have shown ways to intentionally hallucinate the responses of LLMs via adding additional information \citep{freda2023distract}, varying the prompt engineering methods \citep{reliable2023gpt, dress2023}, or providing guidance \citep{cov2024, coni2023}. Unlike the past paradigm, we show that prompting the LLMs with no additional information but simply to make another attempt leads to decreasing the chances of reaching the solution.
    \item \textbf{Suggest a novel method for editing complex free response questions.} Many studies have shown promising results in the LLMs ability to break down complex problems into smaller components \citep{least2023most, jason2022cot} or rewinding fixing code from chain of revisions \citep{code2023chain}. However, we suggest a new direction that allows users to solve complex problems by breaking the original problem into easier sub-components. This eliminates the need for users to find relevant sample problems in order to prompt the LLM accordingly.
    \item \textbf{Provide a prompting method applicable to encouraging learning.} Many variation of Chain-of-Thought (CoT) prompting methods such as \citet{verify2023edit} risk the chance that LLMs generate incorrect answers with no way for users without domain-intensive knowledge to verify. However, our novel prompting method allows users to break down problems into smaller steps until they believe it does not require intensive knowledge to verify. Then, from that point onwards, the users are able to digest the reasoning steps towards the final solution and actually understand the logical flow.
\end{itemize}
\begin{figure*}[t!]
    \centering
    \includegraphics[width=\linewidth]{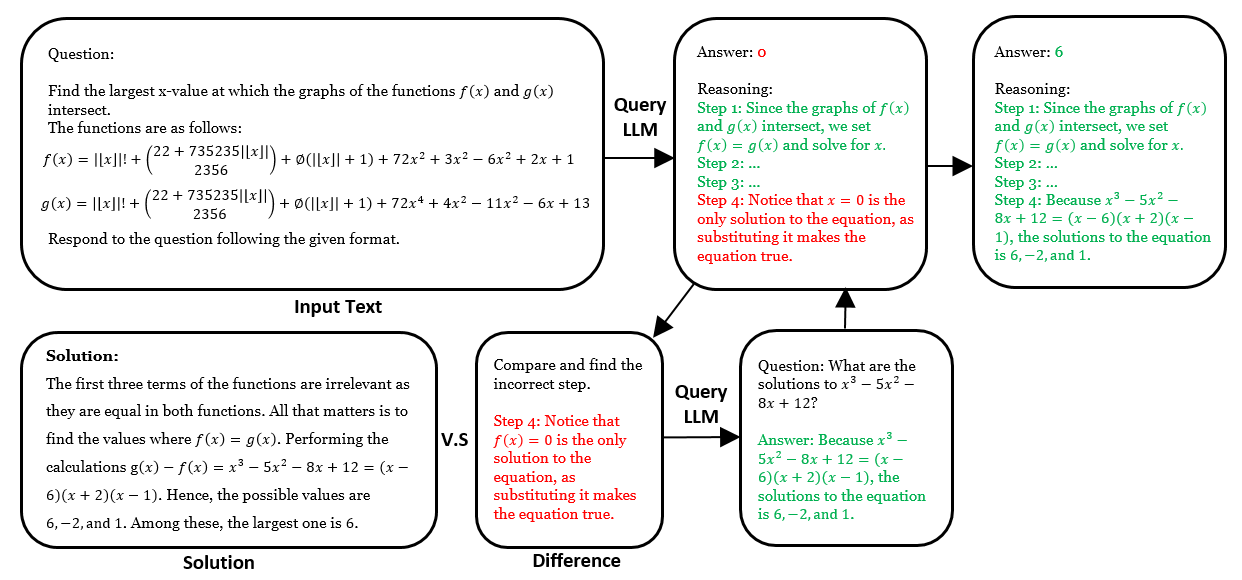}
    \caption{All questions contain definitive solutions as shown above. The R-CoF framework recursively fixes the incorrect reasoning step while having the other steps frozen. The question shown above is from the MATH dataset presented in \citet{hendrycks2021math}.
    }
    \label{fig:problem}
\end{figure*}
\section{Chain-of-Feedback Setting}
In this section, we 1) introduce the background of the problem (i.e. how the CoF setting is a common prompting practice in the real world), 2) share the preliminary insights indicating how meaningless feedback causes more deviation from the solution.

\subsection{Background}
Since the introduction of conversational agents like ChatGPT, one of the very common practices in the real world is to repetitively prompt to request another trial without providing meaningful feedback. Imagine a student struggling to find an answer to a homework problem. The student might decide to prompt LLMs to receive insights. However, realizing that the initial response is incorrect, the student might meaninglessly prompt the LLM to try again only to have the quality of the response get worse. Such behavior of LLMs have only been assumed and yet to be carefully studied by field experts. For this reason, we design a small-scale CoF setting to perform a preliminary experiment that shows insights towards how repetitive prompting without meaningful information drops the quality of the response instead of increasing it.

\subsection{Insights}
To show that LLMs have the tendency to deviate from the answer more with repetitive requests, we performed a preliminary experiment on ChatGPT-3.5 that showed us an insight towards this behavior. We first randomly sample $n$ questions from a selected dataset \citep{hendrycks2021math}. After prompting the LLM to respond and given that the response is incorrect, we provide a meaningless prompt that does not provide much feedback for its previous response (\textit{'Your response is incorrect. Please make another attempt.'}) until it correctly adjusts its previous response or we repeat the process 7 times. For each iteration of response, we calculate the absolute difference between the solution and the generated response.
\begin{algorithm*}
\caption{Recursive Chain-of-Feedback}
\begin{algorithmic}[1]
\Function{RecursiveCoF}{$problem$}
    \State \textbf{Input:} $problem$ that requires multiple steps and is open-ended.
    \State \textbf{Output:} Reasoning steps and final solution to the input.
    \State $initResponse\gets$
    \Call{QueryLLM}{$problem$}

    \If{\Call{IsCorrect}{$initResponse$}}
        \State \Return {$initResponse$}
    \Else
        \State $incorrectStep \gets$ \Call{IdentifyIncorrectStep}{$initResponse$}
        \State $subProblem \gets$
        \Call{IncorrectStepToSubProblem}{$incorrectStep, initResponse$}
        \State $adjustedStep \gets$
        \Call{RecursiveCoF}{$subProblem$}
        \State $adjustedProblem \gets$
        \Call{Replace}{$adjustedStep, initResponse$}
        \State \Return \Call{RecursiveCoF}{$adjustedProblem$}
    \EndIf
\EndFunction
\end{algorithmic}
\label{algo:rcof}
\end{algorithm*}

\subsection{Penalty Rule}
Oftentimes, prompting the LLMs to continuously make another attempt forces it to 'give up'. Common responses that indicate such intentions were those that state 1) there is no solution to the problem when there clearly is, 2) there are infinitely many solutions when there is a unique solution, and 3) repeating the same incorrect answer for multiple times\footnote{We define this number to be three.}. We decide that such responses communicate the intentions of giving up because LLMs are designed to always generate a response. In other words, they inherently do not have the capabilities to communicate their limitations unless explicitly designed to do so \citep{dont2023know}. Therefore, we assume that the list of behaviors above indicates failure to make a genuine attempt.

In order to penalize the LLM for 'giving up', we implement the following simple yet effective function that penalizes the deviation from the actual response: take the maximum deviation thus far, add the penalty term, and make it the deviation for that iteration. If this process continues, the penalty term increases exponentially for every repetition. The intention of the design of this penalizing term is that repeating its previously incorrect response provides less insight to the user than a still incorrect but new attempt. The function is shown for reference:
\[
f(A, R, n) = |A - R| + 10 \times 2^n
\]
where $A$ represents the correct answer, $R$ represents the generated response, and $n$ represents the number of times the LLM has given up thus far. We understand the brevity of the current penalty function and are in the process of constructing a more nuanced one. For now, we seek to communicate the idea that the LLMs should be penalized for 'giving up' to respond.
\section{Recursive Chain-of-Feedback}
As described earlier, a trivial feedback requesting another attempt only forces the language models to deviate more from the target. Similarly, prompting methods like chain-of-thought \citep{jason2022cot} are not the most plausible in our situation because they require a sample question that follows a similar reasoning procedure. Hence, we present a novel method that tackles these issue. For a more detailed execution of R-CoF, refer to Appendix \ref{sec:appendix8.1}.

Unlike normal prompting practices that focus on answering the question as a whole, we propose a novel method that simplifies a complex problem by prompting the LLM to not only respond to the question but also to provide a step-by-step explanation towards the solution. We first query the LLM to provide a response to the problem with reasoning steps. If the initial response is correct, we output that response. Otherwise, we continue. Among those steps, we manually identify the incorrect step. Then, the user prompts a separate language model to solve only that particular step with the same approach. If the LLM is correct, the user incorporates the adjusted step to the original reasoning. If the LLM is incorrect, the user repeats the recursive call until the LLM responds correctly.

\subsection{Benefits}
There are several benefits to R-CoF over existing error-correcting methodologies. First, a complicated problem that an individual cannot even attempt at beginning can be broken down into much simpler steps that can be comprehended. For instance, the problem presented in Figure \ref{fig:problem} might seem intimidating in the beginning, causing a student to not know how to approach the problem at all. Naively asking it to solve the question for them will most likely cause the response to not only fail to reach the correct outcome with the increasing complexity of the input but also provide complicated reasoning that might not seem intuitive enough to comprehend. Nevertheless, by utilizing R-CoF, the problem is broken down into multiple simpler steps that the user can verify the correctness at ease. Then, by adjusting the previously incorrect step and incorporating it to the original lines of reasoning, the LLM will be more likely to be led to the correct direction.

Second, R-CoF does not require a sample question that closely resembles the input question. For example, chain-of-thought prompting \citep{jason2022cot} requires users to include a similar question along with the logical process to the solution so that the LLM can replicate the reasoning. In real-world complex reasoning scenarios, however, finding a similar problem and knowing how to solve it correctly might be equally, if not more, difficult than answering the question of intent. 

\subsection{Issues}
Oftentimes, despite the correct adjustments made by utilizing R-CoF, LLMs fail to incorporate the feedback. For instance, the LLMs will make an acknowledging statement along the lines of \textit{'You are correct. I did make a mistake in Step 4'} and does not take the correct feedback into account when generating a new response. This inherent problem of LLMs are to be studied. We detail such behavior in Appendix \ref{sec:appendix8.2}.
\section{Preliminary Results}
This section outlines the preliminary experimental results of R-CoF on questions that the LLM failed to respond even with repeated trials.
\begin{table}[ht]
\centering 
\begin{tabular}{lccc}
\toprule
Model & Calls & Accuracy & Change \\
\midrule
ChatGPT-3.5 &  & 0/50 & \\
ChatGPT-3.5 & $N$ = 1 & 31/50 & +62\% \\
ChatGPT-3.5 & $N$ = 2 & 37/50 & +12\% \\
\bottomrule
\end{tabular}
\caption{The table reports the accuracy with and without R-CoF. $N$ represents the number of recursive calls}
\label{tab:results}
\end{table}

We test R-CoF on 50 randomly sampled questions from the MATH dataset \citep{hendrycks2021math} that ChatGPT-3.5 fails to respond correctly. Our preliminary results limit the the number of recursive calls to $N=1$ and $N=2$ for brevity, but this number can go as far as the user intends to in the automated version that we wish to develop. Results show that R-CoF accurately corrects 31 of the 50 incorrect questions with one recursive call and 6 additional questions with two recursive calls.
\section{Conclusion}
We perform \textbf{C}hain-\textbf{o}f-\textbf{F}eedback (CoF) to show that meaningless repetitive prompting requesting another trial decreases the chances of the user reaching the correct output. To mitigate the troubles derived from this issue, we present a novel prompt engineering method called \textbf{R}ecursive \textbf{C}hain-\textbf{o}f-\textbf{F}eedback (R-CoF) which 1) divides the question into multiple simple steps, 2) identifies the incorrect steps, and 3) adjusts that particular step on a different setting to ultimately reach the correct solution. We hope that the new prompt engineering method, despite its incompleteness as an ongoing work for now, adds contribution to the idea of correcting errors in LLM reasoning through breaking down complex problems into simpler problems without the need for any other sample data points provided to the language models.
\section{Limitations}
Most of the limitations of this paper are from the nature of the project as an ongoing work. Despite the interesting findings presented in the paper, there exists a lack of quantitative experiments that holds back the technical details of the paper at its current form. The performance of R-CoF also has not been extensively studied enough to compare with existing state of the art prompting methods on different benchmarks other than mathematical reasoning questions. Further, the method has not been tested only on ChatGPT-3.5 and not on other language models. We seek to address and are actively addressing these limitations for the final version of the paper. Finally, R-CoF requires a manual identification of the error in the reasoning. Although we are working towards automating this process, even as is, we believe that the users will be able to discern the reasoning errors because R-CoF breaks down complex problems into smaller components. Finally, the experiments in this paper are performed on public conversational language models, so we cannot guarantee the experimental results are fully reproducible.
\section{Related Work}
\textbf{Prompt Engineering}
Since the introduction of chain-of-thought prompting \citep{jason2022cot}, recent works have proposed prompt engineering methods that increase the reliability of language model responses. \citet{self2023consistent} propose the idea of generating multiple responses on the same question and selecting the answer with the most occurrences, relying on the assumption that the final answer should be the same regardless of the approach. More works suggest that language models are capable of correcting reasoning errors \citep{cov2024, error2023correct, google2023aspire}. Nevertheless, these works all fail to recognize the idea that language models can continuously generate the same incorrect answer as shown in the CoF setting and/or require additional information other than the question itself. R-CoF, on the other hand, effectively tackles these issues, requiring only the question itself as the input.

\noindent\textbf{Error Correcting}
Previous studies propose promising error-correcting techniques \citep{error2023correct, verify2023edit}. Yet, they suffer from focusing excessively on tackling the problem as a whole. \citet{an2023lema} suggests a similar idea as ours in correcting errors in reasoning but is expected to suffer from more complex reasoning tasks. This is as shown in our preliminary experimental results where language models occasionally fail to adjust the incorrect reasoning step. However, our method of R-CoF recursively calls the function to break down the questions until it gets it correct, which makes us predict that the overall accuracy in different benchmarks will be higher.

\bibliography{latex/acl_latex}

\begin{thebibliography}{19}
\expandafter\ifx\csname natexlab\endcsname\relax\def\natexlab#1{#1}\fi

\bibitem[{An et~al.(2023)An, Ma, Lin, Zheng, Lou, and Chen}]{an2023lema}
Shengnan An, Zexiong Ma, Zeqi Lin, Nanning Zheng, Jian-Guang Lou, and Weizhu Chen. 2023.
\newblock \href {http://arxiv.org/abs/2310.20689} {Learning from mistakes makes llm better reasoner}.

\bibitem[{Chen et~al.(2023{\natexlab{a}})Chen, Yoon, Ebrahimi, Arik, Pfister, and Jha}]{google2023aspire}
Jiefeng Chen, Jinsung Yoon, Sayna Ebrahimi, Sercan~O Arik, Tomas Pfister, and Somesh Jha. 2023{\natexlab{a}}.
\newblock Adaptation with self-evaluation to improve selective prediction in llms.
\newblock In \emph{Empirical Methods in Natural Language Processing}.

\bibitem[{Chen et~al.(2023{\natexlab{b}})Chen, Sikka, Cogswell, Ji, and Divakaran}]{dress2023}
Yangyi Chen, Karan Sikka, Michael Cogswell, Heng Ji, and Ajay Divakaran. 2023{\natexlab{b}}.
\newblock Dress: Instructing large vision-language models to align and interact with humans via natural language feedback.
\newblock \emph{arXiv:2311.10081}.

\bibitem[{Dhuliawala et~al.(2024)Dhuliawala, Komeili, Xu, Raileanu, Li, Celikyilmaz, and Weston}]{cov2024}
Shehzaad Dhuliawala, Mojtaba Komeili, Jing Xu, Roberta Raileanu, Xian Li, Asli Celikyilmaz, and Jason Weston. 2024.
\newblock Chain-of-verification reduces hallucination in large language models.
\newblock In \emph{International Conference on Learning Representation}.

\bibitem[{Hendrycks et~al.(2021)Hendrycks, Burns, Kadavath, Arora, Basart, Tang, Song, and Steinhardt}]{hendrycks2021math}
Dan Hendrycks, Collin Burns, Saurav Kadavath, Akul Arora, Steven Basart, Eric Tang, Dawn Song, and Jacob Steinhardt. 2021.
\newblock Measuring mathematical problem solving with the math dataset.
\newblock In \emph{Advances in Neural Information Processing Systems}.

\bibitem[{Huschens et~al.(2023)Huschens, Briesch, Sobania, and Rothlauf}]{doyou2023trust}
Martin Huschens, Martin Briesch, Dominik Sobania, and Franz Rothlauf. 2023.
\newblock Do you trust chatgpt? -- perceived credibility of human and ai-generated content.
\newblock \emph{arXiv preprint arXiv:2309.02524}.

\bibitem[{Le et~al.(2024)Le, Chen, Saha, Gokul, Sahoo, and Joty}]{code2023chain}
Hung Le, Hailin Chen, Amrita Saha, Akash Gokul, Doyen Sahoo, and Shafiq Joty. 2024.
\newblock Codechain: Towards modular code generation through chain of self-revisions with representative sub-modules.
\newblock In \emph{International Conference on Learning Representation}.

\bibitem[{Lei et~al.(2023)Lei, Li, Hu, Wang, Ching, and Kamal}]{coni2023}
Daren Lei, Yaxi Li, Mengya~(Mia) Hu, Vincent Wang, Mingyu abd~Yun, Emily Ching, and Eslam Kamal. 2023.
\newblock Chain of natural language inference for reducing large language model ungrounded hallucinations.
\newblock \emph{arXiv:2310.03951}.

\bibitem[{Mündler et~al.(2023)Mündler, He, Jenko, and Vechev}]{niels2023contradict}
Niels Mündler, Jingxuan He, Slobodan Jenko, and Martin Vechev. 2023.
\newblock Self-contradictory hallucinations of large language models: Evaluation, detection and mitigation.
\newblock In \emph{International Conference on Learning Representation}.

\bibitem[{Ouyang et~al.(2022)Ouyang, Wu, Jiang, Almeida, Wainwright et~al.}]{openai2022}
Long Ouyang, Jeff Wu, Xu~Jiang, Diogo Almeida, Carroll~L. Wainwright, et~al. 2022.
\newblock Training language models to follow instructions with human feedback.
\newblock \emph{arXiv preprint arXiv:2203.02155}.

\bibitem[{Shi et~al.(2023)Shi, Chen, Misra, Scales, Dohan, Chi, Schärli, and Zhou}]{freda2023distract}
Freda Shi, Xinyun Chen, Kanishka Misra, Nathan Scales, David Dohan, Ed~Chi, Nathanael Schärli, and Denny Zhou. 2023.
\newblock Large language models can be easily distracted by irrelevant context.
\newblock In \emph{International Conference on Machine Learning Representations}.

\bibitem[{Si et~al.(2023)Si, Gan, Yang, Wang, Wang, Boyd-Graber, and Wang}]{reliable2023gpt}
Chenglei Si, Zhe Gan, Zhengyuan Yang, Shuohang Wang, Jianfeng Wang, Jordan Boyd-Graber, and Lijuan Wang. 2023.
\newblock Prompting gpt-3 to be reliable.
\newblock In \emph{International Conference on Learning Representation}.

\bibitem[{Tyen et~al.(2023)Tyen, Mansoor, Cărbune, Chen, and Mak}]{error2023correct}
Gladys Tyen, Hassan Mansoor, Victor Cărbune, Peter Chen, and Tony Mak. 2023.
\newblock Llms cannot find reasoning errors, but can correct them!
\newblock \emph{arXiv:2311.08516}.

\bibitem[{Wang et~al.(2023)Wang, Wei, Schuurmans, Le, Chi, Narang, Chowdhery, and Zhou}]{self2023consistent}
Xuezhi Wang, Jason Wei, Dale Schuurmans, Quoc Le, Ed~Chi, Sharan Narang, Aakanksha Chowdhery, and Denny Zhou. 2023.
\newblock Self-consistency improves chain of thought reasoning in language models.
\newblock \emph{arXiv:2203.11171}.

\bibitem[{Wei et~al.(2022)Wei, Wang, Schuurmans, Bosma, Ichter, Xia, Chi, Le, and Zhou}]{jason2022cot}
Jason Wei, Xuezhi Wang, Dale Schuurmans, Maarten Bosma, Brian Ichter, Fei Xia, Ed~Chi, Quoc Le, and Denny Zhou. 2022.
\newblock Chain-of-thought prompting elicits reasoning in large language models.
\newblock In \emph{Advances in Neural Information Processing Systems}.

\bibitem[{Yin et~al.(2023)Yin, Sun, Guo, Wu, Qiu, and Huang}]{dont2023know}
Zhangyue Yin, Qiushi Sun, Qipeng Guo, Jiawen Wu, Xipeng Qiu, and Xuanjing Huang. 2023.
\newblock Do large language models know what they don't know?
\newblock In \emph{Association for Computational Linguistics}.

\bibitem[{Zhang(2023)}]{peter2023advice}
Peter Zhang. 2023.
\newblock Taking advice from chatgpt.
\newblock \emph{arXiv preprint arXiv:2305.11888}.

\bibitem[{Zhao et~al.(2023)Zhao, Li, Joty, Qin, and Bing}]{verify2023edit}
Ruochen Zhao, Xingxuan Li, Shafiq Joty, Chengwei Qin, and Lidong Bing. 2023.
\newblock Verify-and-edit: A knowledge-enhanced chain-of-thought framework.
\newblock In \emph{Association for Computational Linguistics}.

\bibitem[{Zhou et~al.(2023)Zhou, Schärli, Hou, Wei, Scales, Schuurmans, Cui, Bousquet, Le, and Chi}]{least2023most}
Denny Zhou, Nathanael Schärli, Le~Hou, Jason Wei, Nathan Scales, Dale Schuurmans, Claire Cui, Oliver Bousquet, Quoc Le, and Ed~Chi. 2023.
\newblock Least-to-most prompting enables complex reasoning in large language models.
\newblock In \emph{International Conference on Learning Representation}.

\end{thebibliography}

\clearpage
\onecolumn
\section{Appendix}
\subsection{Sample R-CoF Execution}
\label{sec:appendix8.1}
Below is an example execution of R-CoF on a problem from MATH \citep{hendrycks2021math}. We map each step below to Algorithm \ref{algo:rcof} by referencing the appropriate function calls inside the parentheses.\\

\noindent\textbf{Initial User Input (Input)}\\
Respond to the question below with the following format:\\
Reasoning (e.g. Step N...)\\
Question:\\
If $h(x)$ is a function whose domain is $[-8, 8]$, and $g(x) = h(\frac{x}{2})$, then the domain of $g(x)$ is an interval of what width? \\

\noindent\textbf{Initial GPT 3.5 Output (initResponse)} \\
Reasoning:\\
Step 1: The function $g(x)$ is defined as $g(x) = h(\frac{x}{2})$, which means that $g(x)$ takes the input $x$, divides it by $2$, and then applies the function $h$.\\
Step 2: To find the domain of $g(x)$, we need to consider the domain of $h(x)$ and how it is affected by the transformation $\frac{x}{2}$.\\
Step 3: Since the domain of $h(x)$ is $[-8, 8]$, when we divide $x$ by $2$, the values of $x$ will be halved. So, the domain of $g(x)$ will be affected by this transformation. \\
\textcolor{red}{Step 4: Dividing the original domain $[-8, 8]$ by $2$, we get $[-4, 4]$.} \\
Step 5: The width of the interval $[-4, 4]$ is $4 - (-4) = 4 + 4 = 8$.\\
\textcolor{red}{Step 6: Therefore, the domain of $g(x)$ is an interval of width $16$.} \\

\noindent\textbf{Identification of Incorrect Steps (IdentifyIncorrectStep)}\\
We can identify that Steps 4 and 6 are logically flawed. First, Step 4 is wrong because the domain doubles instead of shrinking. Second, Step 6 is wrong because it does not match with the interval calculated in the previous step, Step 5.\\

\noindent\textbf{Translation of Incorrect Step to Question (IncorrectStepToSubProblem)}\\
If the domain of the function $h(\frac{x}{2})$ is $[-8, 8]$, what is the viable range of $x$?\\

\noindent\textbf{Adjustment of Incorrect Step by LLM (RecursiveCoF)}\\
Since the domain of $h(\frac{x}{2})$ is in $[-8, 8]$, it means that $-8 \leq \frac{x}{2} \leq 8$. \\ 
Now we solve for $x$: \\
$-8 \times 2 \leq x \leq 8 \times 2$\\
$-16 \leq x \leq 16$\\
So, $x$ is in $[-16, 16]$\\

\noindent\textbf{Incorporation of Adjusted Step into Original Response (Replace)}\\
In Step 4, since $\frac{x}{2}$ is in $[-8, 8]$, we can solve for $x$, and thus $x$ is in $[-16, 16]$. Can you solve the original question based on this given information?\\

\noindent\textbf{Refinment of Output (RecursiveCoF)}\\
You're correct. Let's adjust the reasoning accordingly:\\
Step 4: Since $\frac{x}{2}$ is in the interval $[-8, 8]$, we can solve for $x$ to be in $[-16, 16]$. Therefore, the domain of $g(x)$ is in $[-16, 16]$.\\
Step 5: To find the width of the interval $[-16, 16]$, we subtract the endpoints $16 - (-16) = 16 + 16 = 32$.\\
\textcolor{blue}{Step 6: Therefore, the domain of $g(x)$ is an interval of width $32$.}

\subsection{Ignorance of Language Models}
\label{sec:appendix8.2}
Below is an example situation in which the language model fails to incorporate the adjusted step into its new reasoning despite acknowledging its mistake.\\

\noindent The context is such that the discriminant has to be calculated for the equation $x^2 + bx + 2 = -2$, but the LLM is only calculating the discriminant for the left hand side of the equation.\\

\noindent\textbf{Incorporation of Adjusted Step into Original Response (Replace)}\\
In Step 3, we need to find the discriminant of the function such that it does not intersect with the horizontal line $y = -2$. Therefore, we should find the discriminant of the equation $x^2 + bx + 2 = -2$. Please solve the original problem with this information.\\

\noindent\textbf{Refinment of Output (RecursiveCoF)}\\
You are correct. Apologies for the oversight. Let's correct that.\\
\noindent Reasoning:\\
\noindent Step 1: To find the values of $b$ for which $-2$ is not in the range of the function $f(x) = x^2 + bx + 2$, we need to find when the quadratic function $x^2 + bx + 2$ does not intersect the horizontal line $y = -2$.\\
\noindent\textcolor{red}{Step 2: We want to ensure that the quadratic function $x^2 + bx + 2$ has no real roots at $y = -2$. For this, we need to consider the related equation $x^2 + bx + 2 = 0$.} \\
... \\
(LLM output continues to give incorrect answer)

\end{document}